\documentclass[a4paper]{article}

\usepackage{INTERSPEECH2021}

\usepackage{subcaption}
\usepackage[dvipsnames]{xcolor}

\usepackage{diagbox}
\usepackage{hyperref}
\usepackage[symbol]{footmisc}

\title{Few-Shot Keyword Spotting in Any Language}

\name{Mark Mazumder$^1$, Colby Banbury$^1$, Josh Meyer$^2$\textsuperscript{\**}, Pete Warden$^3$, Vijay Janapa Reddi$^1$}

\address{$^1$Harvard University, USA\\
  $^2$Coqui, Germany\\
  $^3$Google, USA}
\email{\{markmazumder,cbanbury\}@g.harvard.edu, josh@coqui.ai, petewarden@google.com, vj@eecs.harvard.edu}

\begin{document}

\maketitle

\begingroup\renewcommand{\thefootnote}{\fnsymbol{footnote}}
\footnotetext[1]{Partially conducted while at Artie Inc. and the Mozilla Foundation}
\endgroup
\begin{abstract}

We introduce a few-shot transfer learning method for keyword spotting in any language. Leveraging open speech corpora in nine languages, we automate the extraction of a large multilingual keyword bank and use it to train an embedding model. With just five training examples, we fine-tune the embedding model for keyword spotting and achieve an average $F_1$ score of $0.75$ on keyword classification for 180 new keywords unseen by the embedding model in these nine languages. This embedding model also generalizes to new languages. We achieve an average $F_1$ score of $0.65$ on 5-shot models for 260 keywords sampled across 13 new languages unseen by the embedding model. We investigate streaming accuracy for our 5-shot models in two contexts: keyword spotting and keyword search. Across 440 keywords in 22 languages, we achieve an average streaming keyword spotting accuracy of 87.4\% with a false acceptance rate of 4.3\%, and observe promising initial results on keyword search.

\end{abstract}

\noindent\textbf{Index Terms}: speech recognition, keyword spotting, low-resource languages

\section{Introduction}\label{sec:intro}

Training keyword spotting (KWS) models requires the manual collection and curation of thousands of target samples across a diverse pool of speakers and accents for each keyword of interest \cite{warden2018speech} --- a prohibitive requirement for under-resourced languages. In this paper, we relax the training data for a KWS model to just five training examples in any language.

We train an embedding model on keyword classification using Common Voice's~\cite{ardila2019common} multilingual crowd-sourced speech dataset, by applying forced alignment~\cite{mcauliffe2017montreal} to automatically extract 760 frequent words across nine languages. We then fine-tune this embedding model to classify a target keyword with just five sample utterances, even if the model has never seen the target language before. We evaluate our embedding representation's performance on 440 keywords across 22 languages to demonstrate the generalization of our approach to languages and words previously unseen by the embedding model.

Our contributions are as follows: (1) we achieve promising 5-shot keyword spotting accuracy across 22 languages via a multilingual embedding representation; (2) we show multilingual embeddings improve accuracy and generalize to new languages; (3) we highlight the value of crowd-sourced data in low-resource settings for evaluating classification and streaming accuracy;  and (4) we open-source our code and models and provide a Colab for easy reproduciblity and extension.\footnote{\href{https://github.com/harvard-edge/multilingual_kws}{https://github.com/harvard-edge/multilingual\_kws}}

These contributions lay the groundwork towards a fully automated, rapid time-to-solution pipeline for generating voice-based command interfaces for arbitrary keywords in low-resource languages. Our current aim is to enable a volunteer to record just 5 examples of a target keyword in a zero-resource language and obtain a robust multi-speaker KWS model. Our future efforts will target deployment on low-cost, power-efficient microcontrollers for always-on KWS support.

\begin{figure}
    \centering
    \includegraphics[width=\linewidth]{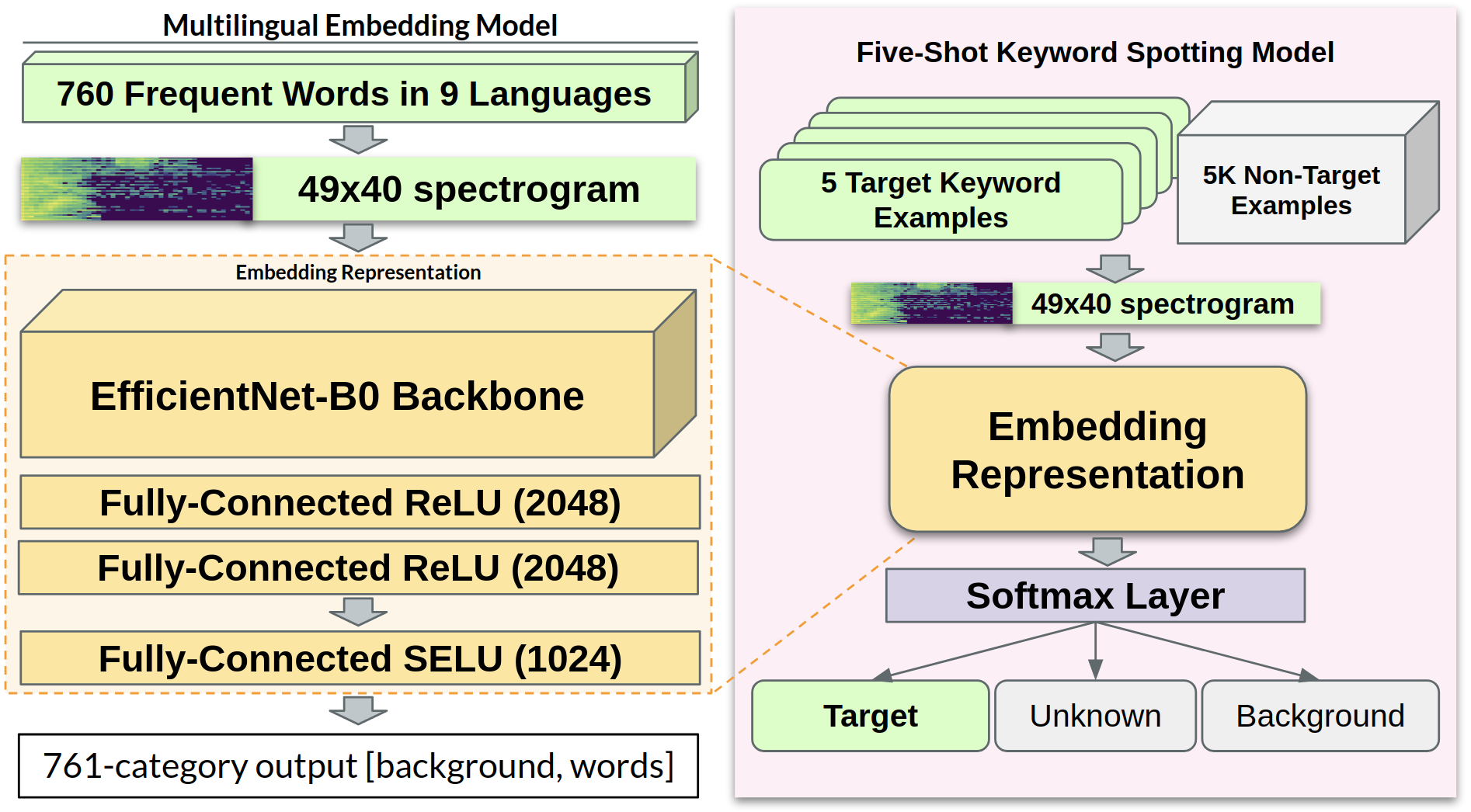}
    \begin{minipage}[t]{.5\linewidth}
    \subcaption{Multilingual embedding model}\label{fig:classifier}
    \end{minipage}%
    \begin{minipage}[t]{.5\linewidth}
    \centering
    \subcaption{5-shot keyword spotting}\label{fig:fewshot}
    \end{minipage}
    \vspace{-10pt}
    \caption{Multilingual Embedding Representation: \textup{(a)} To learn a multilingual embedding for keyword feature extraction, we train a classifier on 760 keywords totaling 1.4M samples in nine languages, and use the output of the penultimate layer of our classifier as a feature vector for arbitrary keywords in any language. \textup{(b)} To train a new KWS model, we fine-tune a 3-category classifier using just 5 target examples and 128 non-target samples from a precomputed ``unknown'' keyword bank. }
    \label{fig:sysarch}
    \vspace{-10pt}
\end{figure}

\section{Related Work}\label{sec:related}

Many approaches to keyword spotting have been proposed. Prior art focused on developing small footprint models for keyword spotting tasks using  deep neural networks \cite{chen2014small}, convolutional neural networks \cite{sainath2015convolutional, zhang2017hello, tang2018deep, banbury2021micronets}, and long short-term memory neural networks \cite{chen2015query}. Similarly, we use a CNN based architecture for our classifier, however, these previous methods require thousands of samples of the target keyword. In contrast, our work only requires 5 keyword samples in a target language, which enables keyword spotting in low resource languages.

For KWS in low-resource languages, existing methods~\cite{Menon2019, menon2018fast} employ multilingual bottleneck features, dynamic time warping, and autoencoders, requiring roughly 30 samples per keyword in addition to a few hours of untranscribed data. In \cite{huang2021querybyexample} the authors demonstrate high KWS accuracy with 3 examples for English and Korean. Bluche et al.~\cite{Bluche2020} shows promising accuracy in English for a zero-shot approach to detecting any keyword, Awasthi et al.~\cite{awasthi2021teaching} evaluates a few-shot embedding in 5 languages, and San et al.~\cite{san2021leveraging} performs spoken term search in 10 languages. Our work utilizes a simpler embedding scheme and considers few-shot performance across a comparatively large number of languages and speakers, using only 5 training examples of a target keyword. While speech synthesis has also shown recent success in KWS~\cite{lin2020training}, we target languages which lack sufficient data for synthesis.

\section{Multilingual Keyword Spotting}

In this section, we describe our system for training a multilingual embedding model, performing transfer learning, and automating the extraction of a large keyword dataset.

\subsection{Multilingual Embedding Model} \label{sec:embedding}
Our network architecture is summarized in Fig.~\ref{fig:classifier}. We repurpose the output of the penultimate layer of a simple keyword classifier as our embedding representation in our few-shot experiments. Our classifier uses TensorFlow Lite Micro's \cite{david2020tensorflow} microfrontend spectrogram \cite{tfmicrofrontend} as 49x40x1 inputs. It contains approximately 11 million parameters and consists of a randomly-initialized EfficientNet-B0 implementation from Keras \cite{kerasb0}, followed by a global average pooling layer, two dense layers of 2048 units with ReLU activations, and a penultimate 1024-unit SELU activation \cite{klambauer2017self} layer, before the classifier's 761-category softmax output. We chose SELU activations for their self-normalizing properties. The classifier is trained on 760 words across nine languages (listed in Table~\ref{tab:embacc}) and 1.4 million samples in total (Sec.~\ref{sec:gen}). We select the most common words in each language and filter by character length of 3 or higher to discard brief words and stop words. Each extraction is padded with silence to one second in length. We also include a background noise category in the output, with 10\% of all training samples consisting solely of noise sampled from background noise examples in Google's Speech Commands dataset \cite{warden2018speech}. Keyword samples are augmented with random 100ms timeshifts, background noise multiplexed at 10\% SNR, and SpecAugment \cite{Park2019}.

\subsection{Few-shot Transfer Learning} \label{sec:xfer} 
For 5-shot transfer learning (Fig. \ref{fig:fewshot}), we use five target samples to fine-tune a 3-class softmax layer (with \textit{target, unknown,} and \textit{background} categories) on the output feature vector of the embedding layers, along with 128 non-target samples drawn from a precomputed bank of 5,000 ``unknown'' utterances in the nine embedding languages. When training KWS models in languages not seen by the embedding model (e.g., in Welsh), non-target samples are still drawn from this bank, i.e., to train a KWS model in Welsh a user would only need to collect 5 target samples of a Welsh keyword, without also needing to collect non-target examples in Welsh. The weights in the embedding layers are frozen when fine-tuning; we only update the softmax layer. Across 256 total training samples, approximately 45\% are in the target category (random augmentations of the five target examples using the same strategy as Sec.~\ref{sec:embedding}), 45\% are negative samples drawn from the precomputed set of non-target words, and 10\% are background noise (Sec.~\ref{subsubsec:res:ile}).

\subsection{Dataset Generation} \label{sec:gen}

Our extracted keywords are entirely sourced from Common Voice \cite{ardila2019common}. We use the Montreal Forced Aligner \cite{mcauliffe2017montreal} to estimate word-level alignments for each \textless{} audio,transcript \textgreater{} pair in Common Voice.\footnote{\href{https://github.com/JRMeyer/common-voice-forced-alignments}{https://github.com/JRMeyer/common-voice-forced-alignments}} We performed forced alignment from a flat start only on the data itself, with no prior acoustic models or external data. We automate keyword extraction using the alignment timings. For our experiments, we extracted 4,383,489 samples across 3,126 keywords in 22 languages.

\section{Experiments}\label{sec:experiments}

We evaluate our capabilities for KWS in multiple languages through classification and streaming accuracy experiments.

\subsection{Classification Accuracy} \label{subsec:exp:class}

We assess classification performance for the embedding model, and for five-shot KWS models evaluated on a large number of extracted target and non-target keywords. 

\subsubsection{Multilingual embeddings trained on extracted keywords} \label{subsubsec:exp:mle}
In order to evaluate the quality of our embedding representation trained on extracted keywords (Sec~\ref{sec:embedding}), we assess the top-one accuracy of the classifier on a validation set of 161,700 samples. Furthermore, we report the validation accuracy for each language to inspect whether it is skewed. We train the embedding model for 94 epochs using the Adam optimizer from Keras with a learning rate of 0.001.

We also inspect the potential domain gap between extracted and manually recorded keywords.  We cross-compare the test accuracy of a \textit{tinyconv} model \cite{tftinyconv} trained on the keyword ``left'' chosen randomly from the Google Speech Commands (GSC) dataset \cite{warden2018speech} and a model trained on keyword extractions of ``left'' from Common Voice English data. After training a \textit{tinyconv} model on Common Voice data, we assess the model's classification performance on GSC data and vice versa.

\subsubsection{Monolingual vs. multilingual embeddings} \label{subsubsec:exp:ile}
We explore the accuracy of a multilingual embedding relative to individual language embedddings, by comparing KWS models fine-tuned on each. We train six monolingual embedding models by selecting 165 frequent words per language and using a penultimate layer width of 192 units. We evaluate KWS models for 20 \textit{out-of-vocabulary} words (i.e., words unseen when training the embedding representation) in the target language. We select up to 2,000 samples per keyword and train a KWS model by fine-tuning on 5 samples. We evaluate classification performance by validating on the remaining 1,995 samples as positive class examples, along with 30,000 samples across 90 non-target words as negative examples. To assess keyword classification accuracy for the multilingual embedding, for each of the nine languages in the embedding, we randomly select 20 target words distinct from the 760 keywords used to train the multilingual representation (Sec.~\ref{sec:embedding}), and evaluate classification performance for 5-shot models against all other positive samples of each keyword and 30,000 negative samples across 90 non-target words. Negative examples are divided evenly between keywords previously used to train the embedding model (these should now be categorized as non-targets by the 5-shot model, and not misclassified as the target), keywords sampled from the bank of unknown samples used when fine-tuning 5-shot models (Sec.~\ref{sec:xfer}), and previously unencountered keywords which are novel to both the embedding and KWS models.

\subsubsection{Out-of-embedding classification} \label{subsubsec:exp:ooec}
We investigate whether the multilingual embedding can be used to perform keyword spotting in languages not seen by the embedding model, i.e., we establish whether a precomputed multilingual embedding can generalize to other languages without collecting additional training data in that language beyond five samples of a target keyword.  We consider classification accuracy across two settings: (1) \textit{out-of-vocabulary} words not previously seen by the multilingual embedding model, but spoken in the languages used to train the embedding, and (2) \textit{out-of-embedding} words in languages unseen when training the multilingual embedding model. %

\subsection{Few-Shot Streaming Accuracy} \label{subsec:exp:stream}

In practice, KWS models operate on a continuous stream of audio, thus we inspect streaming accuracy in two regimes. (1) We concatenate individual words with an average gap of 2 seconds (filled with random background noise), to simulate wake-word or command interaction with a voice assistant. For each KWS model, we evaluate on 10 minutes of audio containing approximately 100 keywords and 100 random non-target words. (2) We search for keywords in continuous spoken audio, by concatenating approximately 20 minutes of full sentences from Common Voice for each keyword under evaluation. Our streaming post-processing approach follows Sec. 7.2 in \cite{warden2018speech}. In each regime, we randomly select out-of-vocabulary and out-of-embedding words across 22 languages for a total of 440 KWS models. Each KWS model is also fine-tuned against two versions of our embedding model. As described in Sec.~\ref{sec:embedding}, our baseline embedding model is trained on 1.4M samples, each padded out to 1 second with silence. We train a second embedding model using the same hyperparameters on 2.8M 1-second samples, where each previous sample now occurs twice in our dataset, first padded with silence as before, and additionally padded with the surrounding audio from the originating Common Voice clip.

\section{Results} \label{sec:results}

We summarize our classification and streaming accuracy evaluations for 5-shot KWS models. We note that in all of our automated KWS evaluations, the five training samples are randomly selected from forced alignment extractions and are not manually inspected beforehand, hence performance for some models will be negatively affected by poor extractions or errors in the original Common Voice recordings.

\begin{table}
\centering
\caption{Classification Accuracy for Multilingual Embedding Model: We show the number of words per language and number of training samples the embedding model was trained on, followed by the number of validation samples and the validation accuracy of the embedding model for each language.}
\label{tab:embacc}
\begin{tabular}{lrrrr}
\toprule
    Language & \# words &  \# train &   \# val &  val acc \\
\midrule
     English &     265 &   518760 &   57640 &    78.95 \\
      German &     152 &   287100 &   31900 &    79.90 \\
      French &     105 &   205920 &   22880 &    79.16 \\
 Kinyarwanda &      68 &   134640 &   14960 &    73.64 \\
     Catalan &      80 &   132660 &   14740 &    87.63 \\
     Persian &      35 &    69300 &    7700 &    85.70 \\
     Spanish &      31 &    61380 &    6820 &    79.65 \\
     Italian &      17 &    31680 &    3520 &    81.16 \\
      Dutch &       7 &    13860 &    1540 &    72.60 \\
\midrule
      Model &    760 &  1455300 &  161700 &    79.81 \\
\bottomrule
\end{tabular}
\end{table}

\begin{table}
\centering
\caption{Domain gap between our extracted keyword dataset (Extracted) and the manually recorded Google Speech Commands (GSC). The row indicates the training dataset for the model; the column indicates the testing dataset. }
\label{tab:gap}
\begin{tabular}{|c|*{2}{c|}}\hline
\backslashbox{Training}{Test}&GSC&Extracted\\\hline
GSC &93.42\%&90.49\% \\\hline
Extracted &78.07\%&92.23\% \\\hline
\end{tabular}
\vspace{-10pt}
\end{table}

\subsection{Classification Accuracy Results} \label{subsec:res:class}

\begin{figure*}
    \centering
    \begin{subfigure}{0.3\textwidth}
        \centering
        \includegraphics[width=\linewidth]{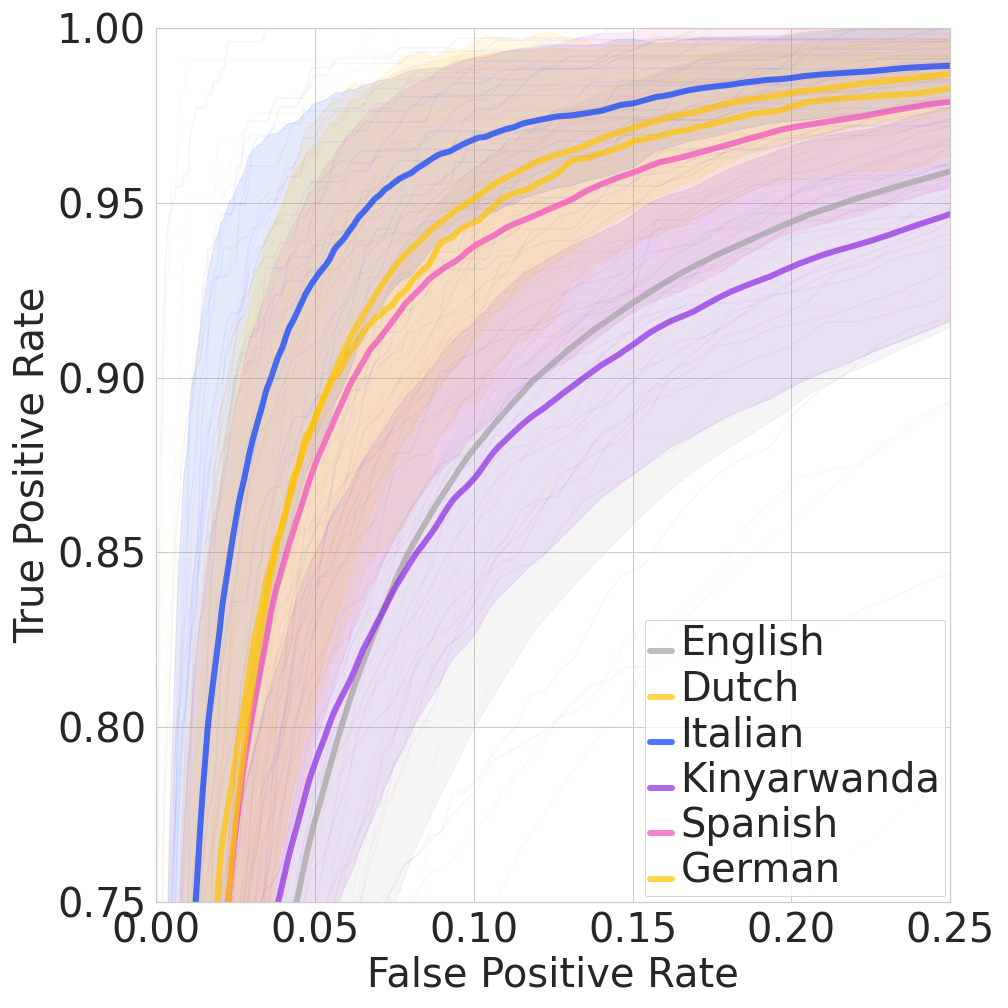}
        \caption{6 Monolingual Embeddings}
        \label{fig:plrc}
    \end{subfigure}
    \hfill
    \begin{subfigure}{0.3\textwidth}
        \centering
        \includegraphics[width=\linewidth]{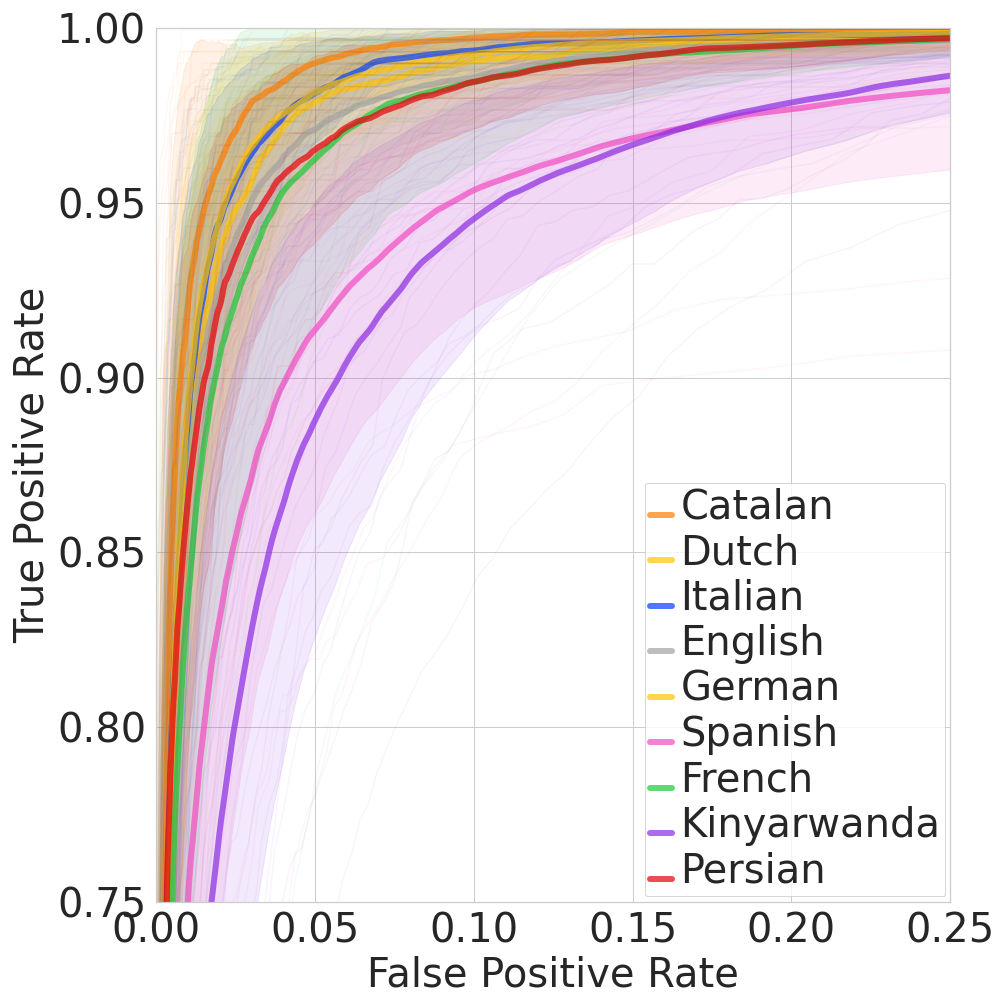}
        \caption{Multilingual Embedding}
        \label{fig:mlrc}
    \end{subfigure}
    \hfill
    \begin{subfigure}{0.3\textwidth}
        \centering
        \includegraphics[width=\linewidth]{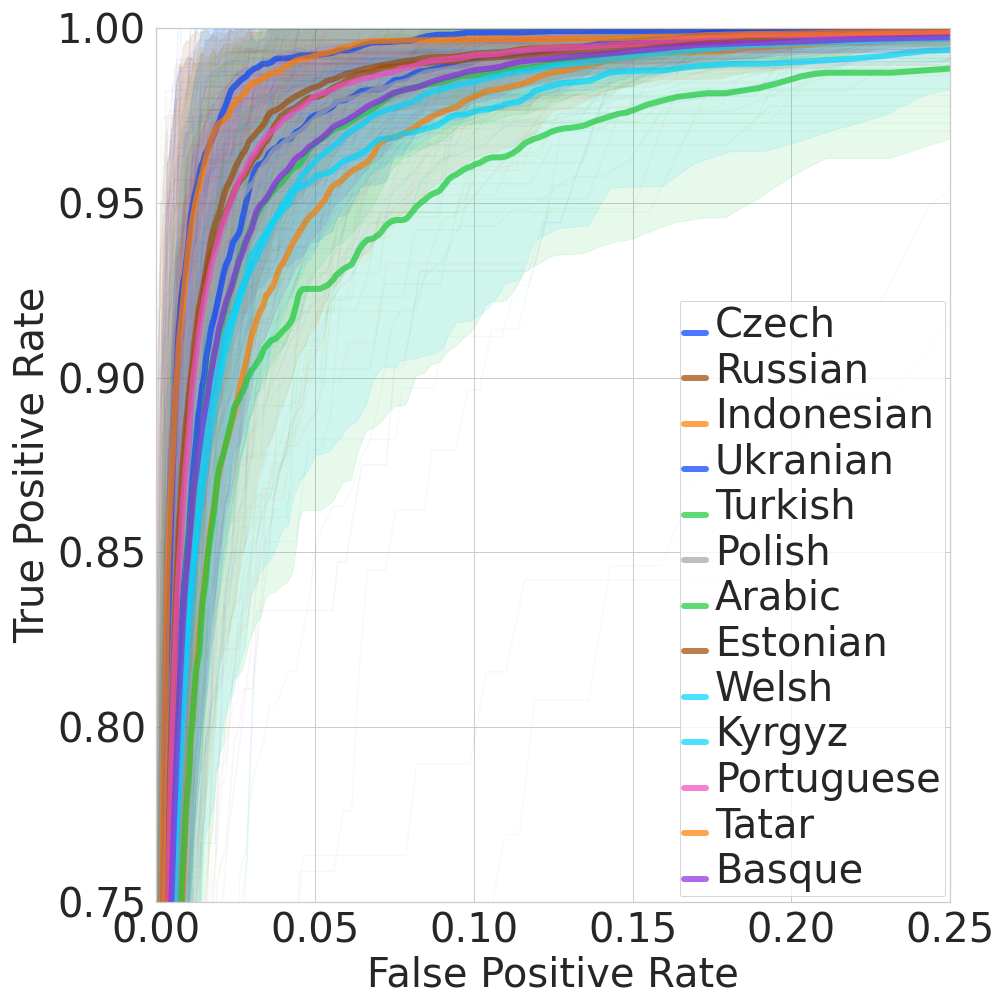}
        \caption{Generalization to New Languages}
        \label{fig:ooec}
    \end{subfigure}
    \caption{5-Shot KWS Classification Accuracy: ROC curves for 5-shot KWS models with 20 randomly selected keywords per language. For each language, the mean is drawn as a bolded curve over the shaded standard deviation (all keywords are shown as a hairline trace). \textup{(a)} 5-shot KWS models using an embedding representation trained per language for six languages. $[$Average $F_1$@$0.8=0.58]$ \textup{(b)} 5-shot models using a multilingual embedding trained on nine languages --- accuracy improves relative to (a). $[$Avg. $F_1$@$0.8=0.75]$ \textup{(c)} 5-shot models using the same multilingual embedding from (b) for random keywords in 13 languages which are out-of-embedding (i.e., which the feature extractor has never encountered), showing that our embedding generalizes to new languages. $[$Avg. $F_1$@$0.8=0.65]$}
    \label{fig:multilang_embedding_classification}
    \vspace{-10pt}
\end{figure*}

\begin{figure}
    \centering
    \begin{subfigure}{0.23\textwidth}
        \centering
        \includegraphics[width=\linewidth]{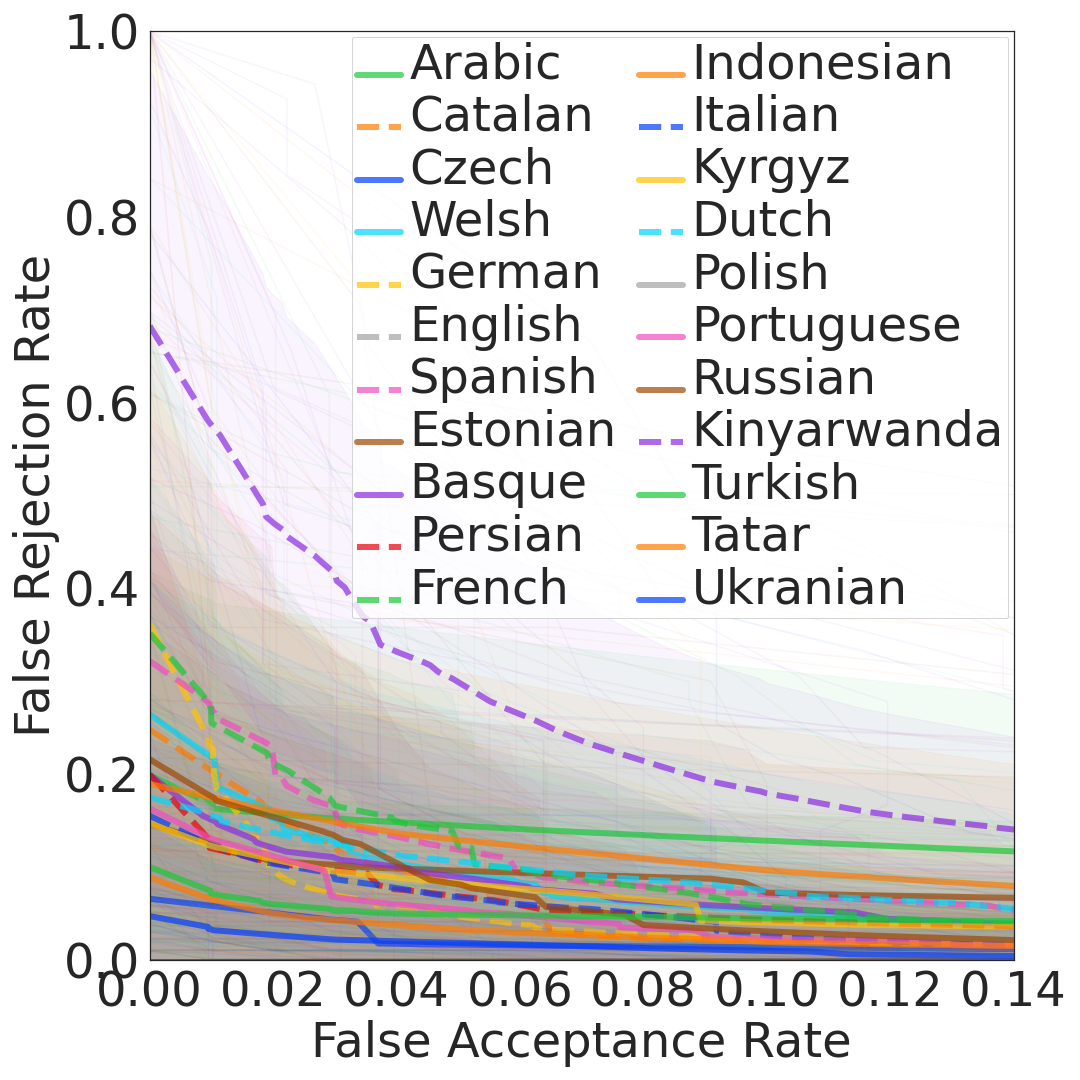}
        \caption{Baseline Keyword Spotting}
        \label{fig:silence_spot}
    \end{subfigure}
    \begin{subfigure}{0.23\textwidth}
        \centering
        \includegraphics[width=\linewidth]{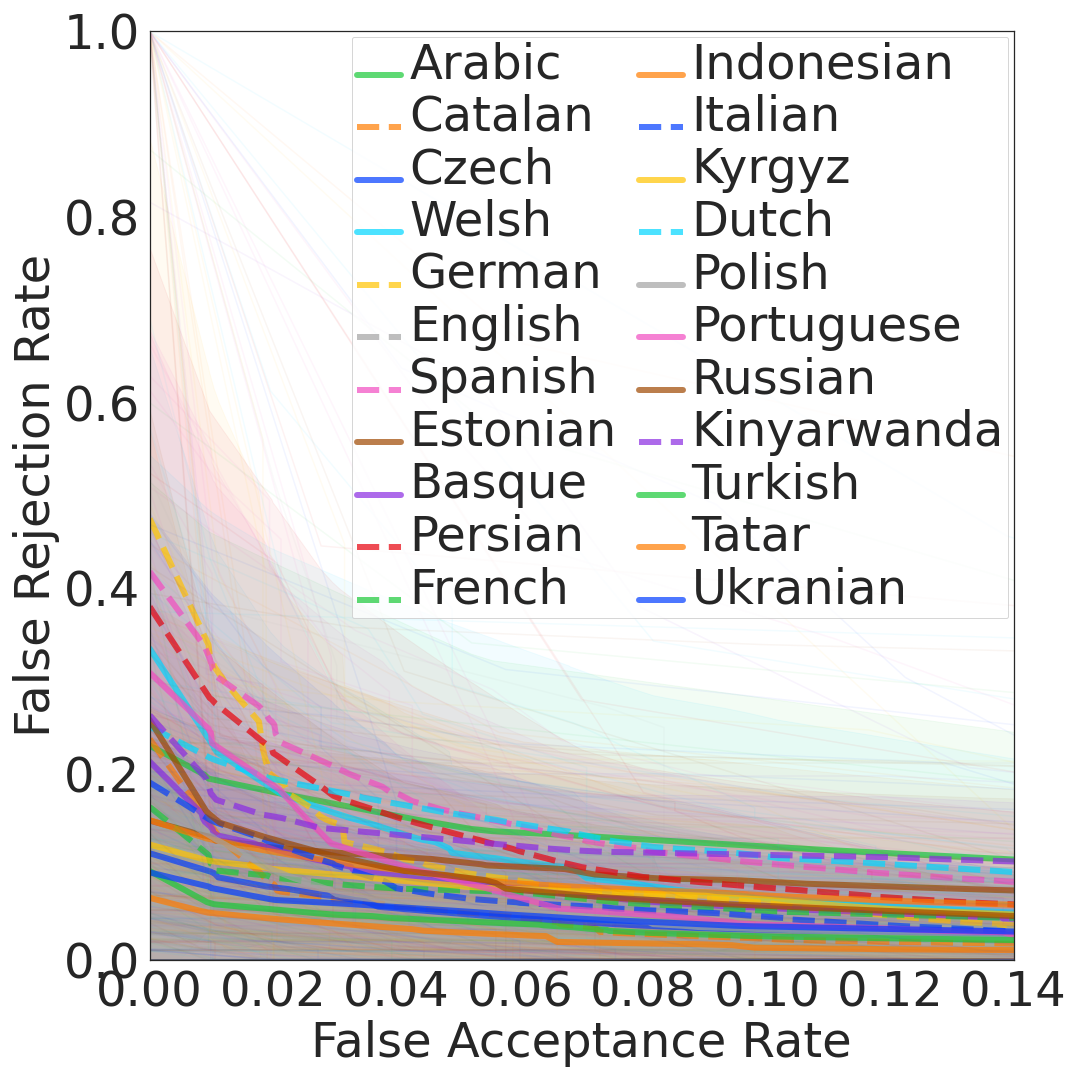}
        \caption{Context Keyword Spotting}
        \label{fig:context_spot}
    \end{subfigure}
    \begin{subfigure}{0.23\textwidth}
        \centering
        \includegraphics[width=\linewidth]{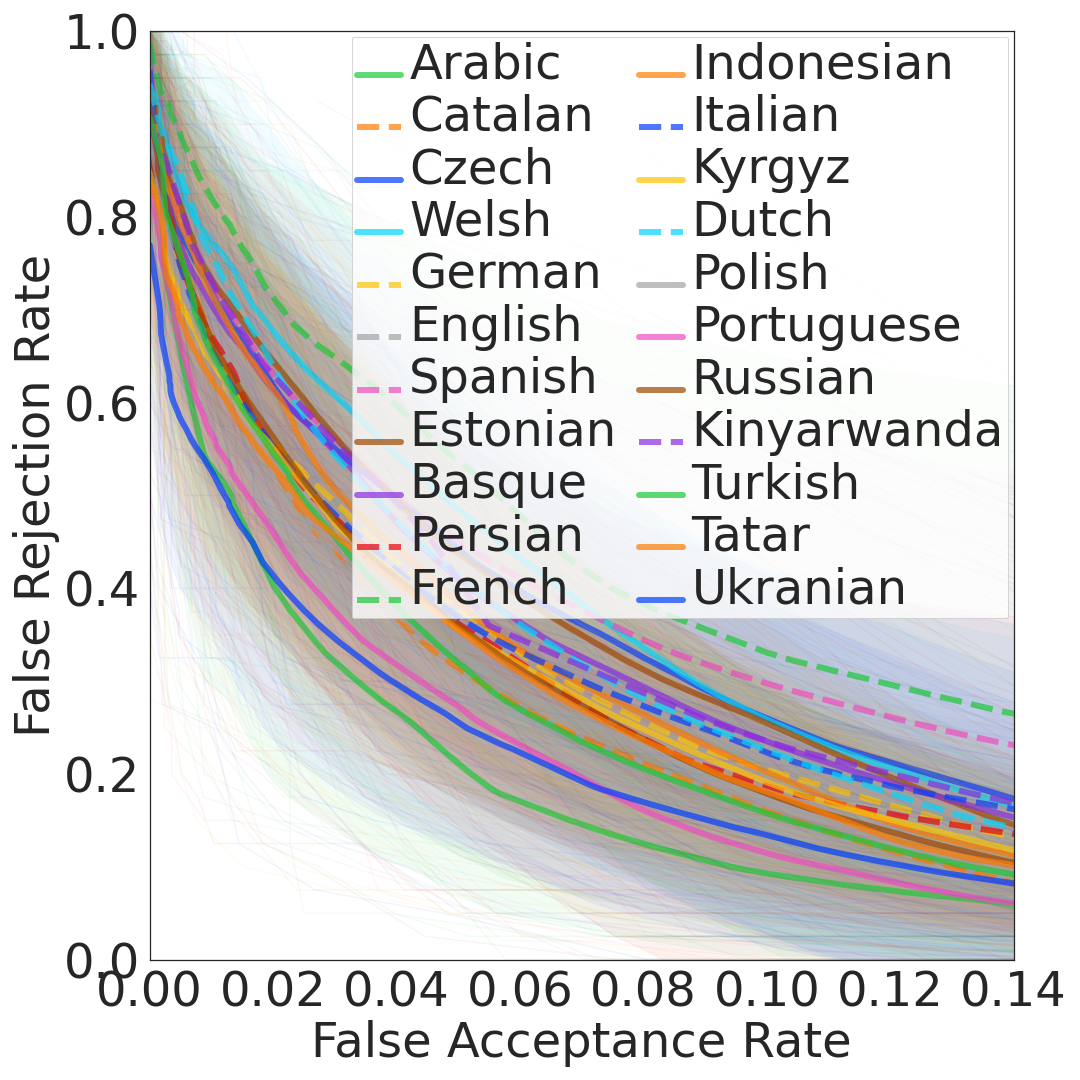}
        \caption{Baseline Keyword Search}
        \label{fig:silence_search}
    \end{subfigure}
    \begin{subfigure}{0.23\textwidth}
        \centering
        \includegraphics[width=\linewidth]{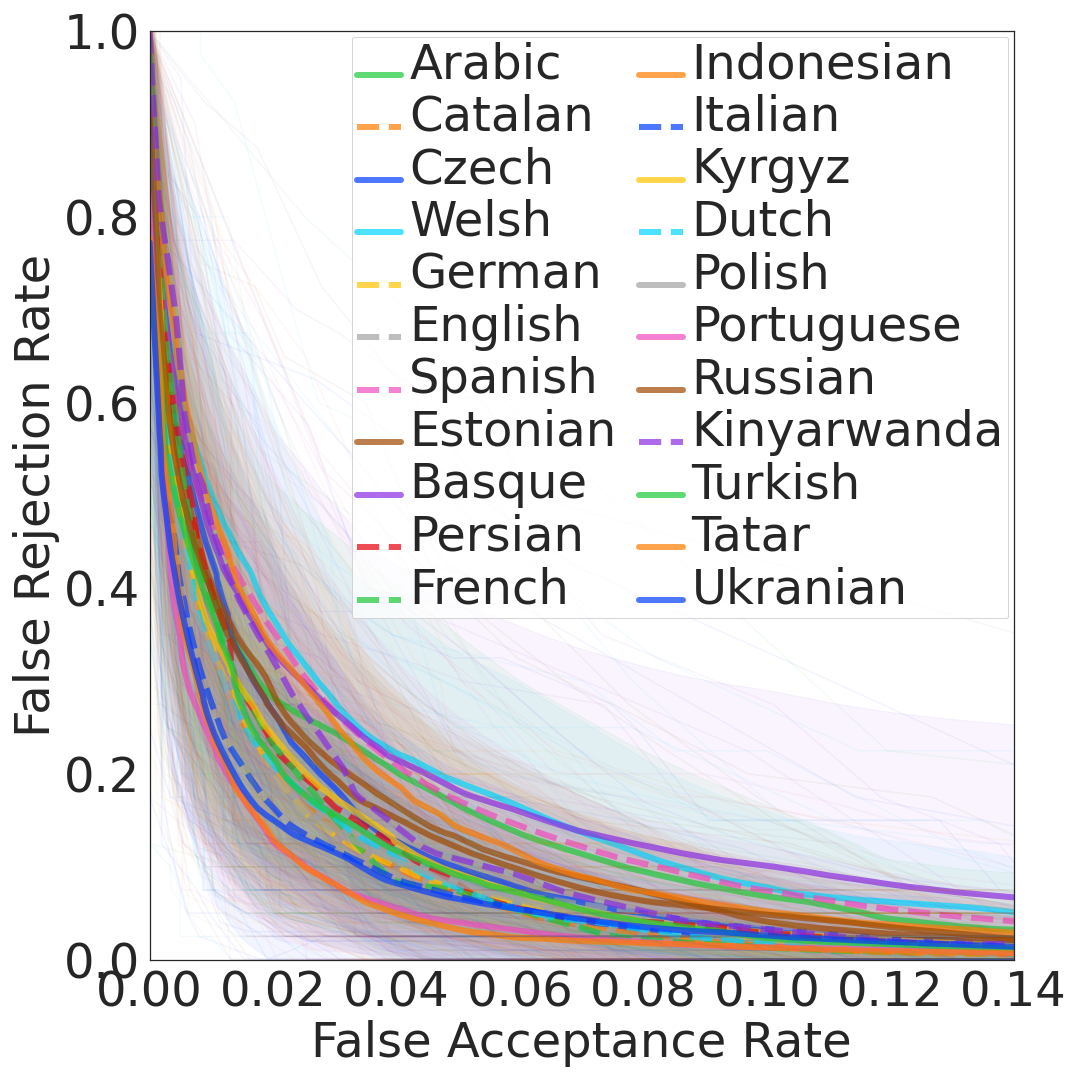}
        \caption{Context Keyword Search}
        \label{fig:context_search}
    \end{subfigure}
    
    \caption{Streaming Accuracy: 5-shot KWS models evaluated on: \textup{(a),(b)} a stream of target and non-target words emulating a wakeword setting, and \textup{(c),(d)} keyword search on a stream of spoken sentences. (b) and (d) demonstrate the improved accuracy from training our embedding representation on both silence- and context-padded samples, over a silence-padded-only baseline (Sec.~\ref{subsec:exp:stream}). At a threshold of 0.8, (b) achieves an average TPR of 87.4\% and FPR of 4.3\% over 22 languages: 9 languages which have been seen by the embedding (dashed lines) and 13 out-of-embedding languages (solid lines). (d) shows our keyword search capabilities using the embedding model trained with contextual audio, achieving an average @0.8 TPR of 77.2\% and FPR of 2.3\%, a significant improvement over (c) the baseline embedding representation.} 
    \label{fig:streaming}
    \vspace{-10pt}
\end{figure}

We evaluate (1) the training performance of our multilingual embedding model, (2) improvements to KWS classification accuracy when using a multilingual embedding representation versus monolingual embeddings, and (3) KWS classification accuracy in languages outside of the multilingual embedding.

\subsubsection{Embedding model accuracy} \label{subsubsec:res:emb}

Table~\ref{tab:embacc} summarizes the accuracy of our multilingual embedding model (Sec.~\ref{sec:embedding}) on a validation set for each of the nine languages used in training. The embedding model achieves an overall classification accuracy of 79.81\%.

As a microbenchmark to compare extracted samples versus manually recorded samples, Table~\ref{tab:gap} reports our cross comparison of two \textit{tinyconv} models trained on Common Voice extractions and GSC data for the keyword ``left.'' The model trained on Common Voice extractions performs worse on the GSC test set than vice versa, dropping to roughly 78\%. Our future work will explore the causes of this apparent domain gap and seek to reduce it. Additionally, we will compare our method against state-of-the-art models trained on GSC.

\subsubsection{Monolingual vs multilingual embedding results} \label{subsubsec:res:ile}

We find that using the multilingual embedding in Sec.~\ref{subsubsec:res:emb} improves KWS classification accuracy versus monolingual embeddings. Fig.~\ref{fig:multilang_embedding_classification} shows receiver operating characteristic (ROC) curves where the false positive rate is visualized against the true positive rate as the threshold for the target keyword is varied for each 5-shot KWS model (Fig.~\ref{fig:fewshot}). 20 previously unseen words were randomly chosen as targets for each language evaluated. Each model was fine-tuned on 256 samples with a batch size of 64 (Sec~\ref{sec:xfer}).  The number of samples and batch size were chosen empirically via a hyperparameter sweep.

Fig.~\ref{fig:plrc} shows classification accuracy for 5-shot models using six monolingual embeddings. Fig.~\ref{fig:mlrc} shows the classification accuracy achieved on 5-shot models using the multilingual embedding representation.  
Accuracy improves for all languages by using the multilingual representation. With an empirically chosen threshold of 0.8, the average (unweighted) $F_1$ score across all KWS models increases from 0.58 to 0.75. For example, despite there being no additional data in Kinyarwanda between Fig.~\ref{fig:plrc} and Fig.~\ref{fig:mlrc}, classification accuracy for Kinyarwanda still improves. As suggested in~\cite{Menon2019, Hermann2018}, generalizable features across languages may benefit each language's accuracy.

\subsubsection{Out-of-embedding classification results} \label{subsubsec:res:ooec}

Fig.~\ref{fig:ooec} depicts classification accuracy for 20 common words chosen randomly from each of 13 languages unobserved when training the embedding model. Accuracy remains high for the majority of these languages, with an average $F_1$ score of 0.65 at a threshold of 0.8, suggesting the multilingual embedding model generalizes beyond the languages seen when training it.

\vspace{-2pt}
\subsection{Five-Shot Streaming Accuracy Results} \label{subsec:res:stream}

Fig.~\ref{fig:streaming} reports streaming accuracy as the false acceptance rate vs. the false rejection rate per instance of true positives in each streaming regime (Sec.~\ref{subsec:exp:stream}). Figs.~\ref{fig:silence_spot} and~\ref{fig:context_spot} report our wakeword emulation results on 440 10-minute audio segments, with no performance penalty observed when using the silence- and context-padded embedding representation versus the baseline. Many KWS models exhibit high precision and recall despite having only one to five speakers in each training set. Figs.~\ref{fig:silence_search} and~\ref{fig:context_search} depict our performance on keyword search in 440 20-minute segments containing full sentences of spoken audio. Training the embedding on both silence- and context-padded samples results in significantly higher accuracy (Fig.~\ref{fig:context_search}) over the baseline (Fig.~\ref{fig:silence_search}). Training on keywords with surrounding audio context improves the embedding representation's ability to identify keywords among coarticulation effects in speech.
\vspace{-2pt}

\section{Conclusions}
We demonstrate 5-shot KWS for arbitrary keywords in 22 languages, using an embedding representation pre-trained on an automatically generated dataset. Our embedding generalizes beyond the nine languages it was trained on, and in future work we will investigate support for languages in which word-level alignment is impractical. By training the embedding on keywords with surrounding audio context, we achieve promising accuracy on keyword search in continuous speech. We utilize a simple fine-tuning scheme for few-shot learning, and continuation studies will apply natural extensions from the literature (e.g., \cite{snell2017prototypical}). We will also pursue a smaller embedding representation via knowledge distillation for on-device deployment.

\section{Acknowledgements}
We thank Raziel Alvarez, Greg Diamos, Daniel Galvez, Coleman Hooper, Max Lam, Tejas Prabhune, Jiahong Yuan, MLCommons and MLCommons Research for helpful discussions.

\bibliographystyle{IEEEtran}

\bibliography{multilingual_kws}

\end{document}